\documentclass[aps, letter, floatfix, nofootinbib, twocolumn, superscriptaddress, pre]{revtex4-1}

\usepackage[utf8]{inputenc}
\usepackage{epsfig}
\usepackage[T1]{fontenc}
\usepackage[english]{babel}
\usepackage[table]{xcolor}
\usepackage{t1enc}
\usepackage{graphicx}
\usepackage{amssymb}
\usepackage{amsmath}
\usepackage{relsize}
\usepackage{overpic}
\usepackage{bm}
\usepackage{hyperref}
\usepackage{times}

\usepackage[normalem]{ulem}

\newcommand{\avg}[1]{{\left<#1\right>}}

\newcommand{\dd}{\mathrm{d}}

\usepackage{mathtools}
\def\multiset#1#2{\ensuremath{\left(\kern-.3em\left(\genfrac{}{}{0pt}{}{#1}{#2}\right)\kern-.3em\right)}}

\usepackage{amsmath}

\usepackage{verbatim}
\usepackage{overpic}
\usepackage{booktabs}
\usepackage{placeins}

\newcommand{\cl}{\mathcal{C}}
\newcommand{\m}{\mathcal{M}}
\newcommand{\bb}{\bm{b}}
\newcommand{\A}{\bm{A}}
\newcommand{\Ao}{\bm{A^O}}
\newcommand{\dA}{\delta\A}
\newcommand{\G}{\bm{G}}

\begin{document}

\title{Consistencies and inconsistencies between model selection and link prediction in networks}

\author{Toni Vallès-Català}
\affiliation{Departament d’Enginyeria Química, Universitat Rovira i
Virgili, 43007 Tarragona, Catalonia, Spain}
\author{Tiago P. Peixoto}
\email{t.peixoto@bath.ac.uk}
\affiliation{Department of Mathematical Sciences and Centre for Networks
and Collective Behaviour, University of Bath, Claverton Down, Bath BA2
7AY, United Kingdom}
\affiliation{ISI Foundation, Via Alassio 11/c, 10126 Torino, Italy}
\author{Marta Sales-Pardo}
\affiliation{Departament d’Enginyeria Química, Universitat Rovira i
Virgili, 43007 Tarragona, Catalonia, Spain}
\author{Roger Guimerà}
\affiliation{Departament d’Enginyeria Química, Universitat Rovira i
Virgili, 43007 Tarragona, Catalonia, Spain}
\affiliation{Institució Catalana de Recerca i Estudis Avançats (ICREA),
Barcelona 08010, Catalonia, Spain}


\begin{abstract}
  A principled approach to understand network structures is to formulate
  generative models. Given a collection of models, however, an
  outstanding key task is to determine which one provides a more
  accurate description of the network at hand, discounting statistical
  fluctuations. This problem can be approached using two principled
  criteria that at first may seem equivalent: selecting the most
  plausible model in terms of its posterior probability; or selecting
  the model with the highest predictive performance in terms of
  identifying missing links. Here we show that while these two
  approaches yield consistent results in most of cases, there are also
  notable instances where they do not, that is, where the most plausible
  model is not the most predictive. We show that in the latter case the
  improvement of predictive performance can in fact lead to overfitting
  both in artificial and empirical settings. Furthermore, we show that,
  in general, the predictive performance is higher when we average over
  collections of models that are individually less plausible, than when
  we consider only the single most plausible model.
\end{abstract}

\maketitle

\section{Introduction}

Real-world complex systems display nontrivial interaction structures. A
principled approach to understand these network structures (and the
processes that give rise to them) is to formulate generative models and
infer their parameters from data.  Unfortunately, for any single
empirical network, an unlimited number of models can in principle be
formulated. Therefore, we need robust and well-founded approaches to
compare models and choose the most appropriate one. Specifically, we
need approaches that can identify parsimonious models that avoid both
\emph{overfitting} --- when purely stochastic fluctuations are mistakenly
incorporated into the structure of overly complicated models --- and
\emph{underfitting} --- when we mistake statistically significant
properties of a network for noise, and wrongly select a model that is
too simplistic.

Despite the importance and intricacies of model selection for network
data, the problem has not been studied systematically.  For years,
network models have been compared based on their ability to reproduce
certain topological features, such as the clustering coefficient, the
degree distribution or the community structure. However, such approaches
are not rigorous and are prone to overfitting, since one can always
design complicated enough models that reproduce any of these properties
with arbitrary precision, but that fail to generalize.

Because of this limitation, it is now becoming common to rely on
model-selection approaches that are better suited to strike a balance
between over and underfitting. These approaches can be either supervised
or unsupervised~\cite{murphy_machine_2012}. In supervised model
selection, we prefer the model with the best capacity to generalize from
the data and predict missing
observations~\cite{clauset_hierarchical_2008,guimera_missing_2009,guimera_predicting_2012}.
In unsupervised model selection, we prefer the model with the highest
probability given the data, which can also be interpreted as the model
that most compresses the
network~\cite{rosvall_information-theoretic_2007,peixoto_parsimonious_2013,come_model_2015,peixoto_nonparametric_2017}.

Both approaches aim to find the most parsimonious model, which captures
all the structure in the data without incorporating any of the
noise. Because of this, one would intuitively expect these two criteria
to agree, especially for asymptotically large networks. Indeed, for much
simpler types of (non-network) models, the consistency of both
approaches has been rigorously shown in specific asymptotic
limits~\cite{shao_linear_1993,
shao_asymptotic_1997,arlot_survey_2010}. However, their implementations
are quite different and, in practice, it is not yet understood in what
regimes discrepancies should be expected.

Here, we discuss the probabilistic foundations of supervised and
unsupervised model selection, and make a systematic comparison between
both approaches using variations of the stochastic block
model~\cite{holland_stochastic_1983}. We show that the two criteria tend to agree, that is, that the most
predictive model tends to be the one that most compresses the
data. Crucially, however, we show that it is possible to construct
networks where both approaches differ, even in the infinite size limit, and the supervised approach leads to overfitting. In fact, this
non-intuitive discrepancy is also observed in some real networks, albeit for a minority of cases.

Moreover, we find that, although in practice the most predictive model
is often the one that most compresses the data, the reverse is not true --- the most accurate link predictions are not given by the most
compressive model but by an average over less compressive
ones. Remarkably, for all the networks and models we study, this
improvement in predictive power is larger than the improvement obtained using more sophisticated models; or, in other words, averaging over samples of even the simplest models is often more predictive than the single most sophisticated model.

\section{Probabilistic framework and stochastic block model classes}\label{sec:sbm}

\subsection{Probabilistic framework}

Probabilistically, the model selection task consists in finding the
model $\m$ that is most likely to have generated a given network with
adjacency matrix $\A $, that is, the model that maximizes
$P(\m|\A)$. This probability is the Bayesian posterior
\begin{equation}
  P(\m|\A) =
  \frac{P(\A |\m)P(\m)}{P(\A)}  \; ,
  \label{eq:bayes}
\end{equation}
where $P(\A )$ does not depend on the model and thus typically plays no
role in model selection, $P(\A |\m)$ is the marginal likelihood, and
$P(\m)$ is the model prior. Since the model typically has some
parameters $\xi$, the marginal likelihood is obtained by marginalizing
over them\footnote{Note that integrating over the parameters is not a
methodological choice, but rather the probabilistically correct
calculation.}
\begin{equation}
  P(\A |\m) =  \int P(\A |\m, \xi)P(\xi|\m)\,\dd\xi \; .
  \label{eq:margl}
\end{equation}
The priors $P(\m)$ and $P(\xi|\m)$ encode our degree of \emph{a priori}
knowledge about the plausibility of the model (and its parameters), and should
be chosen based on previous experience and general expectations about
the data.\footnote{A common, but somewhat misguided, criticism of the full probabilistic approach
is precisely the need to specify these priors. Note, however, that
alternative maximum likelihood approaches are simply equivalent to:
(i) assuming that the priors are uniform; (ii) approximating the
integral over the parameters as $P(A|\m)\approx P(A|\m, \xi_\m^*)$,
where $\xi_\m^*$ is the maximum likelihood estimator of $\m$'s
parameters, that is, the value of the parameters that contribute the
most to the integral in Eq.~\ref{eq:margl}.} We discuss this issue in
more detail below.

\subsection{Classes of stochastic block models}

Given the general probabilistic framework outlined above, we next
specify the models we consider, including the priors. Although our
arguments are general, here we focus on the family of stochastic block
models (SBM)~\cite{holland_stochastic_1983,nowicki_estimation_2001,guimera_missing_2009}, which are analytically
tractable and expressive enough to enable us to investigate the issues we are
interested in.  In particular, we consider four {\em model classes}
within the SBM family that are defined by the SBM variation {\em and}
the choice of priors. We describe these classes below. For simplicity and without loss of generality, in
what follows we assume that the networks under consideration are
multigraphs where parallel links between nodes are allowed.

We consider two SBM variations: the \emph{traditional}
SBM~\cite{holland_stochastic_1983} and the \emph{degree-corrected}
SBM~\cite{karrer_stochastic_2011,reichardt_interplay_2011,morup_learning_2009}.
The traditional SBM assumes that each node belongs to one (and only one)
group, and that the tendency of two nodes $i$ and $j$ to form links
depends only on their group memberships, $b_i$ and $b_j$. In particular,
the rate at which $i$ and $j$ form links is $\lambda_{b_i b_j}$, which
gives an overall likelihood
\begin{equation}\label{eq:poisson_sbm}
  P(\A |\bm{\lambda},\bb ) = \prod_{i<j}\frac{\lambda_{b_ib_j}^{A_{ij}}e^{-\lambda_{b_ib_j}}}{A_{ij}!} \;.
\end{equation}
Here $\A $ is the adjacency matrix of the network, $\bb $ is the vector
of group memberships, and $\bm{\lambda}$ is the matrix of group-to-group
connectivity rates.

Traditional SBMs generate groups whose nodes have a similar number of
links, which is potentially an unrealistic assumption given that node
degrees are often broadly distributed in networks. To account for this
observation, Karrer and Newman proposed the degree-corrected
SBM~\cite{karrer_stochastic_2011}. Specifically, they added to the model
a propensity $\theta_i$ of each node $i$ to establish links, so that the
likelihood reads,
\begin{equation}\label{eq:poisson_dcsbm}
  P(\A |\bm{\theta},\bm{\lambda},\bb ) = \prod_{i<j}\frac{(\theta_i\theta_j\lambda_{b_ib_j})^{A_{ij}}e^{-\theta_i\theta_j\lambda_{b_ib_j}}}{A_{ij}!} \;.
\end{equation}
Within this formulation, $\theta_i$ is proportional to  $i$'s expected degree and
can be different for nodes in the same group, allowing this model
to accommodate arbitrary degree sequences within groups.

Given either one of these model likelihoods, the marginal in
Eq.~\ref{eq:margl} is obtained by integrating over their parameters,
with the exception of the partition $\bb$, which we leave as part of
the model specification $\m$. For the degree-corrected model we
have
\begin{equation}\label{eq:dcsbm_marginal}
  P(\A |\bb) = \int P(\A |\bm{\theta},\bm{\lambda},\bb )P(\bm{\theta}|\bb)P(\bm{\lambda}|\bb)\,\dd\bm{\lambda}\,\dd\bm{\theta},
\end{equation}
and analogously for the traditional variant. In order to compute the
marginal likelihood and the final posterior of Eq.~\ref{eq:bayes}, we
need to specify the priors $P(\bm{\theta}|\bb)$, $P(\bm{\lambda}|\bb)$
and $P(\bb)$ for the parameters $\bm{\theta}$, $\bm{\lambda}$, and the
partitions $\bb$ of the nodes into groups, respectively.  In the absence of previous
experience, we typically rely on the so-called noninformative priors,
which ascribe the same probability to all allowed parameter
values. However, for SBMs this assumption imposes a ``resolution limit''
to the maximum number of groups that can be inferred to scale as
$\sqrt{N}$, where $N$ is the number of
nodes~\cite{peixoto_parsimonious_2013}. A solution to this issue is to
replace the noninformative prior by a sequence of nested priors that
represent the structure of the network at different scales via a nested
sequence of SBMs~\cite{peixoto_hierarchical_2014}. This \emph{nested}
SBM reduces the resolution limit to $N/\log N$ without introducing any
bias towards a specific mixing pattern. Since the noninformative version
of the model is a special case of the nested one, the latter is expected
in general to produce better fits, since it alleviates one source of
underfitting.

In this work, we consider the four {\em model classes} obtained from
combining the two SBM variations (traditional and degree-corrected) with
the two choices for model priors (noninformative and nested). We refer
to a model $\m=(\bb,\cl)$ as the combination of model class $\cl$ and a
node partition $\bb$~\footnote{We note that this definition differs from
choices made in part of the literature
(e.g. Refs.~\cite{decelle_asymptotic_2011,yan_model_2014}), where the
model is considered as $\m=(\lambda^*,\theta^*,\cl)$, where $\lambda^*$
and $\theta^*$ are maximum-likelihood point estimates of the parameters,
and one sums over all possible partitions $\bb$. Although such decisions
on what to call a ``model'' are largely arbitrary, the one used here
yields \emph{regularized} approaches, where the dimension of the model
(e.g. number of groups and hierarchy depth) are determined from the data
{\it a posteriori}. The definition used in
Refs.~\cite{decelle_asymptotic_2011,yan_model_2014} presumes not only
that the model size is known {\it a priori}, but also that it is
sufficiently small compared to the data, i.e. the average group size
tends to infinity --- something that cannot be guaranteed, and is
unlikely to be true in most empirical networks. Furthermore, making
point estimates of $\lambda$ are in general problematic, as they require
initial guesses that are sufficiently close to the optimum
value~\cite{kawamoto_algorithmic_2018}. The definition used here,
therefore, allows for a more consistent comparison between the
supervised and unsupervised approaches, that does not rely on such
assumptions and is free of some technical limitations. Note also that
the meaning of the word ``model'' used here refers to the underlying
data generating process, not to the posterior probability of partition
labels. In the parametric case, the latter can be mapped to a
generalized Potts model~\cite{decelle_asymptotic_2011}, but this is not
the terminology we use.\label{footnote:pedandic}}.  Therefore, in what
follows, by model selection we mean selection of both the model class
and the optimal partition within that model class.  We use the
parametrization and priors presented in
Ref.~\cite{peixoto_nonparametric_2017}, as well as the inference
algorithm described there.

\section{Supervised and unsupervised model selection}\label{sec:model-selection}

As we mentioned earlier, we are interested in contrasting two approaches
for model selection on network data: (i) the unsupervised approach,
where models are chosen according to their plausibility given the data;
(ii) the supervised approach, where models are chosen according to their
capacity to predict missing links in the network. In what follows we
describe both approaches in more depth.

\subsection{Unsupervised model selection using the posterior probability and minimum description length}

The probabilistic framework previously outlined provides a natural
criterion to select the best model for any particular network. Indeed,
if we wish to compare two specific models $(\bb_1, \cl_1)$ and $(\bb_2,
\cl_2)$, this can be done by computing the ratio $\Lambda$ between their
respective posterior probabilities in the joint model space comprising all models $(\bb,\cl_1)$ and $(\bb,\cl_2)$
\begin{equation}
  \label{eq:lambda}
  \Lambda = \frac{P(\bb_1,\cl_1|\A )}{P(\bb _2,\cl_2|\A )}
  = \frac{P(\A |\bb_1,\cl_1)P(\bb_1|\cl_1)P(\cl_1)}{P(\A |\bb _2,\cl_2)P(\bb_2|\cl_2)P(\cl_2)} \;,
\end{equation}
and when we are a priori agnostic about model classes (that is, $P(\cl_1) = P(\cl_2) = 1/2$)
\begin{equation}
  \Lambda = \frac{P(\A |\bb_1,\cl_1)P(\bb_1|\cl_1)}{P(\A |\bb _2,\cl_2)P(\bb_2|\cl_2)} \;.
\end{equation}
Here, the marginal likelihoods $P(\A |\bb_i, \cl_i)$ are computed
according to Eq.~\ref{eq:margl} and the priors are set for each model
class as described in the previous section.  Then, $\Lambda > 1$ means
that the evidence in the data favors $(\bb _1, \cl_1)$ over $(\bb _2,
\cl_2)$ (and vice versa), and the magnitude of $\Lambda$ gives the
degree of confidence in the decision~\cite{jeffreys_theory_1998}.

This criterion is entirely equivalent to the so-called minimum
description length approach (MDL)~\cite{grunwald_minimum_2007}. This is
easily seen by noting that the description length $\Sigma(\A,\bb; \cl)$ is
defined as \footnote{It is also common to define an ``energy''
$\mathcal{H}(\bb, \cl)$ such that $P(\A |\bb ,\cl)P(\bb|\cl) = \exp
\left[-\mathcal{H}(\bb, \cl)\right]$~\cite{guimera_missing_2009}. This
energy only differs from the description length by a multiplicative
factor. Note also that, for the models classes we consider here, the prior
over partition is independent of the model class, and thus $P(\bb|\cl) = P(\bb)$.}
\begin{equation}
  P(\A |\bb ,\cl)P(\bb|\cl) = 2^{-\Sigma(\A, \bb; \cl)},
\end{equation}
where
\begin{equation}
  \Sigma(\A, \bb; \cl) = -\log_2P(\A |\bb,\cl) - \log_2P(\bb|\cl)
\end{equation}
is the asymptotic amount of bits necessary to encode the data
(e.g. using Huffmann's prefix algorithm) in two stages, by first
encoding the partitions $\bb$, and then the data $\A$, constrained by
the knowledge of $\bb$. From this we
have
\begin{equation}
  \log_2\Lambda = \Sigma(\A, \bb_2; \cl_2)-\Sigma(\A, \bb_1; \cl_1) \; .
\end{equation}
Therefore, choosing the model that is most plausible given the data is
equivalent to choosing the model with the minimum description length
(which can be calculated exactly for the four model classes described in
the previous section~\cite{peixoto_nonparametric_2017}), that is, the
model that most compresses the data.\footnote{Note that $\Lambda$ is defined in Eq.~\ref{eq:lambda} in terms of the model posteriors in the space comprising {\em both} model classes $\cl_1$ and $\cl_2$, and that the ratio is different if one uses, incorrectly, the model posteriors calculated in the model spaces containing a single model class. By contrast, the description length is the same in the joint and separate spaces, except for an irrelevant additive constant $\log_2 P(\cl_i) = 1$ that affects all $\cl_1$ and $\cl_2$ models equally. This makes the description length particularly attractive for model selection, and is a consequence of the fact that the number of bits needed to describe the network is a physical property---when two model spaces $\cl_1$ and $\cl_2$ are joined, the network is described exactly as in the separate spaces except for an extra bit necessary to specify whether we are dealing with the $\cl_1$ or the $\cl_2$ subspace.}
This interpretation also gives an intuitive explanation to why this
criterion avoids under- and overfitting --- either if noise is
incorporated into the model or if it misses any regularity in the data
it will result in an increase of the description length\footnote{Note
that if we are interested in making a statement about an entire model
class (as we define here, see footnote~\ref{footnote:pedandic}), rather
than a specific partition, we need to compute the probability summed
over all partitions, i.e. $ P(\A |\cl) = \sum_{\bb}P(\A |\bb
,\cl)P(\bb|\cl)$. See Ref.~\cite{peixoto_nonparametric_2017} for more
details and Ref.~\cite{valles-catala_multilayer_2016} for an example.}.

\subsection{Supervised model selection using link/non-link prediction}

As discussed above, the quality of a model can also be evaluated based
on its predictive power and, in particular, its performance at
identifying which of the observed non-links in a network are most likely
to actually correspond to links that have been mistakenly left out of
the observation (or conversely, which links are in fact non-links that
were spuriously introduced).\footnote{Since a single sample of our model
comprises an entire network $\A$, one could argue that the more
canonical formulation of the supervised scenario would be to consider a
set of different networks, with the same number of nodes and presumed to
be sampled from the same model, which are then divided into training and
validation sets, that are used to fit the model and evaluate its
predictive power, respectively. However this situation is rarely
encountered in practice, as we have typically access to only a single
instance of a network.}  This task is known as link (or non-link)
prediction.

To give the problem of link prediction a probabilistic treatment
\cite{clauset_hierarchical_2008, guimera_missing_2009} consistent with
the notation above, we need some extra definitions. We denote as $\Ao$
the adjacency matrix of the observed network (with some entries
missing), and the set of missing entries as an additional matrix $\dA$,
such that the complete matrix is $\Ao\cup\dA$. Crucially, within this
formalism $\Ao$ can either represent a complete matrix, e.g. with the
missing edges representing evidence of absence (and therefore being
equivalent to non-edges), or an incomplete matrix where the missing
edges are unobserved, i.e.  represent the absence of evidence and are
therefore different from non-edges. The only requirement is that the
complete matrix $\Ao\cup\dA$ is indeed complete, i.e. it represents a
definite statement on every edge and non-edge, which holds for the two
scenarios above.

The central assumptions we make are that the complete network
$\Ao\cup\dA$ has been generated using some class $\mathcal{C}$ of the
SBM, and that the set of missing entries $\dA$ has been chosen from some
uniform distribution among all possibilities. Based only on these two
assumptions, and independently of the internal structure of the model
used, the probability of missing entries given the observed network and
model class can be computed exactly as (Appendix~\ref{app:link_pred})
\begin{equation}\label{eq:linkpred_cond}
  P(\dA|\Ao,\cl) \propto \sum_{\bb} \frac{P(\Ao\cup\dA|\bb,\cl)}{P(\Ao|\bb,\cl)} P(\bb|\Ao,\cl) \,,
\end{equation}
up to a unimportant normalization constant. In the expression above,
$P(\Ao\cup\dA|\bb,\cl)$ and $P(\Ao|\bb,\cl)$ are the marginal
likelihoods of the complete and observed networks, respectively, and
$P(\bb|\Ao,\cl)$ is the probability of a partition given the observed
network $\Ao$ and the model class $\cl$. Thus,
Eq.~\ref{eq:linkpred_cond} can be computed in practice by sampling
partitions from this distribution using MCMC, and averaging the ratio of
marginal likelihoods. We note that for $P(\Ao|\bb,\cl)$ and
$P(\bb|\Ao,\cl)$ we may consider the missing edges (non-edges) either as
non-edges (edges) or unobserved, without any change at all to resulting
distribution $P(\dA|\Ao,\cl)$, as the different choices only change the
auxiliary weights in the importance sampling. We return to this weighted
average in Sec.~\ref{sec:averaging}, but for the purpose of model
selection we can use the single-point approximation
\begin{align}\label{eq:linkpred_mdl}
  P(\dA | \Ao, \cl) &\approx \frac{P(\Ao\cup\dA|\bb^*,\cl)}{P(\Ao|\bb^*,\cl)} P(\bb^*|\Ao,\cl),\nonumber\\
  &= 2^{-\Delta\Sigma(\bb^*,\cl)} P(\bb^*|\Ao, \cl)
\end{align}
with $\Delta\Sigma(\bb,\cl) =
{\Sigma(\Ao\cup\dA, \bb; \cl) - \Sigma(\Ao, \bb; \cl)}$ being the
difference in description length between the network with the missing
entries added and the network without them, and where
\begin{equation}\label{eq:bstar}
  \bb ^* = \underset{\bb }{\operatorname{argmax}}\, P(\bb|\Ao,\cl)
\end{equation}
is the partition that most contributes to the posterior distribution,
that is, the most plausible partition given the observed network or,
equivalently, the partition that most compresses the observation. Note
that although Eq.~\ref{eq:linkpred_cond} is true in general,
Eq.~\ref{eq:linkpred_mdl} can only be expected to be a good approximation if
the number of entries in $\dA$ is much smaller than in $\Ao$.

Based on this, the predictive power of a model can be quantified by
analyzing its ability to identify missing links or non-links. Indeed,
for an observed network for which we know that some true links (or
non-links) have been removed, we consider each of these false negatives
as an instance of $\delta \A$ and compute their $P(\dA| \Ao,\cl)$. Then,
we compare these values with the same quantity obtained for true
negative links (non-links) that do not exist in the original network. We
measure the AUC (``area under the curve''), that is defined as the
frequency with which a false negative (a removed link or non-link) has a
predictive probability higher than a true negative (a nonexistent link
or non-link); the most predictive model is the one that yields the
highest AUC.

\section{Comparison of unsupervised and supervised model selection}

Having defined our unsupervised and supervised model selection
approaches, we next demonstrate that, perhaps counter-intuitively, both
approaches do not necessarily yield the same results. In other words, we
demonstrate that the most predictive model is not necessarily the most
plausible one or, equivalently, the one that most compresses the data,
even for infinitely large networks. We illustrate this fact with a set
of synthetic networks and then we discuss the results we find for real
networks.

\subsection{Inconsistency for some simple synthetic networks}
\label{sec:artificial}

Here, we describe a case in which unsupervised model selection and
supervised model selection based on link prediction are not
consistent. We focus on the removal of links, instead of non-links, but
our arguments are also valid in that case, and also when both links and
non-links are removed simultaneously. A more precise discussion of this
case, with explicit calculations, is given in Appendix~\ref{app:auc}.

Consider an ensemble of networks with $B$ groups, such that the number
of links within each group is exactly $e_{\text{in}}$ and the number of
links between any pair of distinct groups is exactly
$e_{\text{out}}<e_{\text{in}}$. Other than this, the degrees of individual nodes are
not fixed, so networks are drawn from the traditional SBM.

\begin{figure}
  \begin{tabular}{cc}
    \begin{overpic}[width=.49\columnwidth]{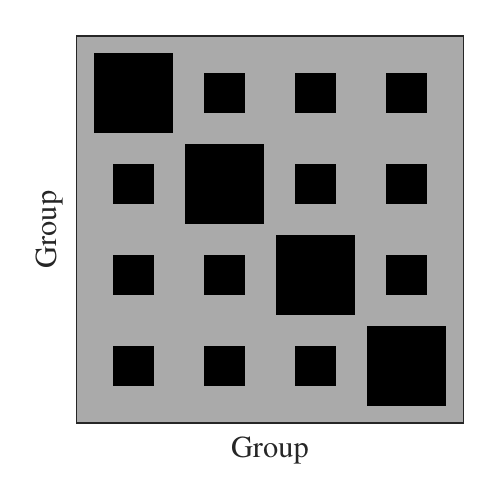}
      \put(0,5){(a)}
    \end{overpic}&
    \begin{overpic}[width=.49\columnwidth]{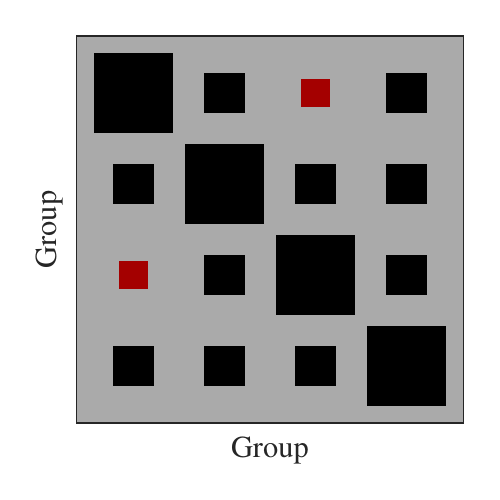}
      \put(0,5){(b)}
    \end{overpic}
  \end{tabular} \caption{(a) Number of edges between groups in a
  synthetic network, before the removal of edges, with $e_{\text{in}}$
  edges in the diagonal and $e_{\text{out}}$ in the off-diagonal,
  represented as squares. (b) The same as in (a), but after a single
  edge has been removed, in the position shown in red (grey). The
  modified entry has $e_{\text{out}}-1$ edges, which will cause the
  predictive likelihood in this position to be lower than for all the
  other entries.\label{fig:loo-diagram}}
\end{figure}

If one removes one inter-group link (between, say, groups $g_1$ and
$g_2$), point-estimate link prediction assuming a traditional SBM will
assign a probability proportional to $e_{\text{out}}$ to all pairs of
nodes between groups $(g_i, g_j) \neq (g_1, g_2)$, and a probability
proportional to $(e_{\text{out}}-1)$ to all pairs between groups $(g_1,
g_2)$, including the one we actually removed (see
Fig.~\ref{fig:loo-diagram}). Therefore, the AUC for this link will be
very low (in fact, lower than 0.5) because most non-links in the network
will have a higher probability of existing than the removed link. As a
matter of fact, one can show that for a large enough number of groups
$B$, the AUC obtained for the complete set of leave-one-out experiments
(i.e. removing one link at a time) will be lower than 0.5 for a broad
range of $e_{\text{in}}$ and $e_{\text{out}}$ (see Eq.~\ref{eq:auc_pp}).

On the other hand, single point-estimate link prediction using
Eq.~\ref{eq:linkpred_mdl} with the degree-corrected SBM will
``absorb'' the missing link into the parameters and assign all
non-observed inter-group links the same approximate probability, thus
providing higher AUCs (see Figs.~\ref{fig:loo}
and~\ref{fig:overfit}). Still, the most parsimonious model in this case
is the traditional SBM and, consistently, the description length is
shorter for that model (because the extra parameters introduced to model
node degrees in the degree-corrected models overfit the data).

Note that the reason why link prediction fails to select the model with
lowest description length in this case is not the lack of statistical
evidence, but rather that the model itself --- and not the data --- is
sensitive to perturbations: A minimal change to one of the
$\lambda_{rs}$ values downgrades the likelihood of the removed edges
with respect to all other edges of the same type that would otherwise
have the exact same probability. Hence, this example illustrates how in
some cases predictive performance (at least when measured by the AUC)
can to some extent reflect inherent properties of a model, rather than
its ability to fit the data.

We emphasize that this scenario is robust with respect to variations of
the types of perturbation done to the network. In particular, if we
remove a non-link instead of a link, we have a symmetric version the
same problem --- the non-link removed will have a lower probability for
precisely the same reason as a removed link. If we consider the removed
link/non-link as an unobserved ``blank'' in the adjacency matrix, as
opposed to its opposite value, this also yields the exact same result,
since our final probabilities only depend on the completed network.

The only situation where one could expect an asymptotic consistency to
be observed is when instead of a single entry of $\A$ we remove a finite
fraction of them at random --- involving links and non-links
indiscriminately. In this situation, we could expect entries between all
pairs of groups to be equally affected on average. However, any
particular set of perturbations would invariably include fluctuations
among group pairs that would yield a similar effect than the one
described here, since the most likely completed network would almost
never be the fully symmetric one in Fig.~\ref{fig:loo-diagram}a.

\subsection{Typical consistency in real networks}\label{sec:empirical}

\begin{figure*}
  \begin{tabular}{cc}
    \begin{overpic}[width=\columnwidth]{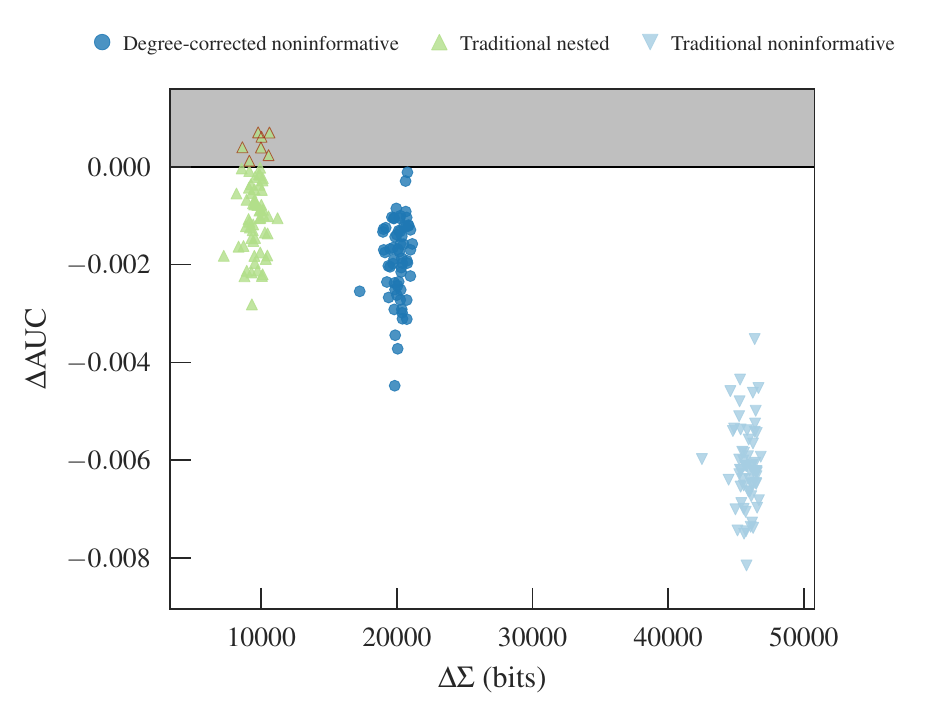}
      \put(0,10){(a)}
    \end{overpic}&
    \begin{overpic}[width=\columnwidth]{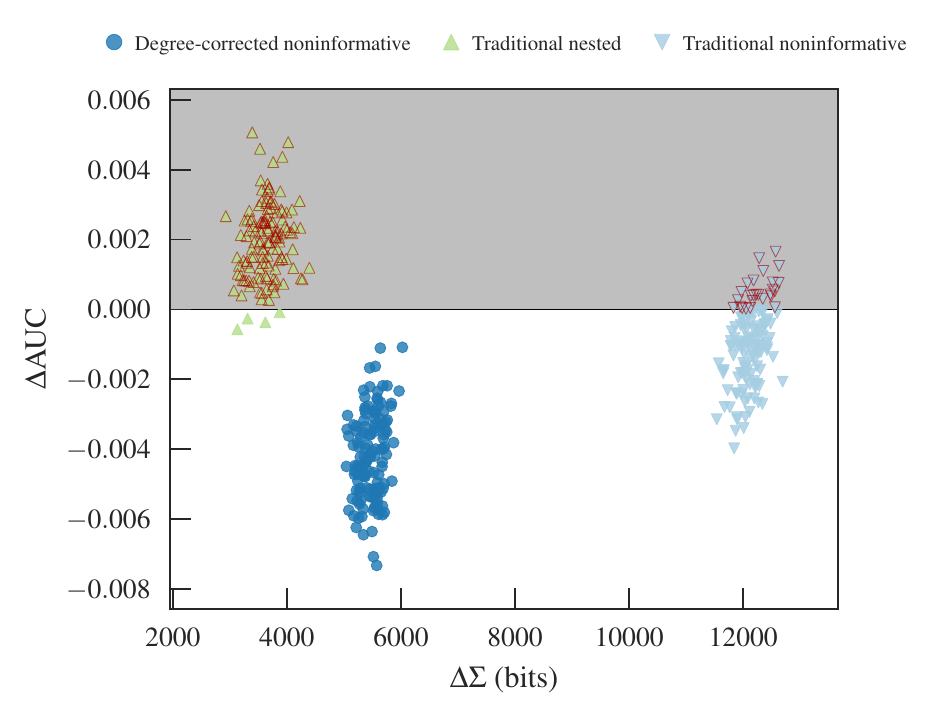}
      \put(0,10){(b)}
    \end{overpic}
  \end{tabular} \caption{Consistency and discrepancy between model selection
  approaches in empirical networks. Comparison between AUC values and the description length
  $\Sigma$, after the removal of a fraction of $f=0.05$ of the
  edges for two empirical networks:
  (a) Global airports; (b) Movielens movie ratings.
  We show results obtained by comparing the degree-corrected nested SBM
  with the three remaining variants, as shown in the legend: traditional
  SBM with noninformative priors, degree-corrected with noninformative
  priors, and traditional with nested priors. Each point corresponds to
  a single instance of the removed edges, and the difference
  $\Delta\text{AUC}$ in AUC and $\Delta\Sigma$ in description length are
  with respect to the degree-corrected nested SBM. Points in the grey
  region represent inconsistent results in which the model with the
  smallest description length (in both cases, the degree-corrected
  nested model) has smaller AUCs (lower predictive
  power).\label{fig:comp}}
\end{figure*}

Given that supervised and unsupervised model selection are not
necessarily consistent, the question is then whether they are consistent
{\em in practice}, that is, in real networks. To answer this question,
we have performed a systematic analysis of the predictive performance of
the four SBM classes on empirical networks (Table~\ref{tab:data}), and
analyzed it \emph{vis a vis} their description lengths.

We observe that, often, supervised and unsupervised model selection are
consistent, meaning the most plausible and most compressive model is
also the most predictive. This is the case, for example, for the air
transportation network (Fig.~\ref{fig:comp}a), for which the best model
overall is the nested degree-corrected SBM, with all the others
displaying both a higher description length (that is, lower
plausibility) and lower AUC values (that is, lower
predictability). However, we also observe a few situations where the
most compressive and most plausible model has an inferior predictive
performance than some of the alternatives. For example, in
Fig.~\ref{fig:comp}b we show the results for the Movielens network of
user-film
ratings~\cite{kunegis_konect:_2013,godoy-lorite_accurate_2016}. For this
network, the nested degree-corrected SBM is the most compressive model,
but the nested traditional SBM provides more accurate predictions of
missing links.

\begin{figure}
  \includegraphics[width=\columnwidth,trim=.25cm 0 .32cm 0,clip]{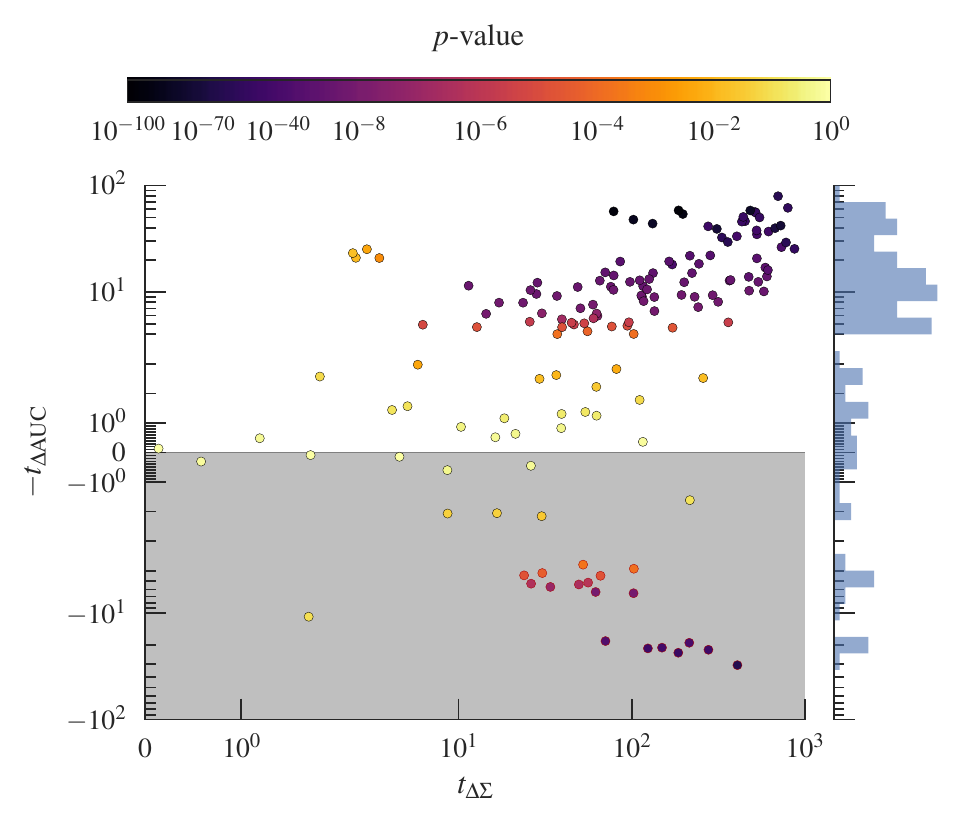}

  \caption{Significance of the discrepancy between model selection
  approaches. For each pair of model classes and for all of the
  empirical networks in table~\ref{tab:data}, we show the t-statistic
  (Eq.~\ref{eq:ttest}) for the difference in AUC and the description
  length $\Sigma$. Results correspond to 30-150 edge removal experiments
  of a fraction $f=0.05$ of the edges (we observe similar results for
  different fractions of removed edges). The number of repetitions for each network is different because for larger networks it takes a longer time to get the results but, at the same time, the fluctuations
between results are smaller. In general, we use as many repetitions as necessary to get reasonable error bars in all
our estimates. Colors (gray tones) show the associated
  $p$-value. Points in the top-half of the figure indicate a consistency
  between both model selection criteria: The model with the smallest
  description length also yields the largest AUC. For points in the
  bottom-half (grey), the comparisons are inconsistent. On the right
  axis, we show a histogram of the $t_{\Delta\text{AUC}}$ values,
  showing that a majority of comparisons are
  consistent.\label{fig:auc_mdl}}
\end{figure}
To quantify how frequent such discrepancies are, we systematically
compare the compressiveness and the predictive power of all model
classes on all networks (Fig.~\ref{fig:auc_mdl}). For each network and
each model class, we generate a noisy observation $\Ao$ by removing a
fraction $f=0.05$ of the links and obtain the optimal partition
$\bb=\bb^*$ as in Eq.~\ref{eq:bstar}. We use this optimal partition to
compute the description length $\Sigma(\bb^*, \cl)$ and the AUC for the
prediction of the missing links. We repeat this operation 30 to 150
times for each real network, and for each set of missing edges and each pair
of models we compute the difference in description length and AUC,
$\Delta\Sigma$ and $\Delta\text{AUC}$. For each model pair, we have a
population of such values, which we use to compute the $t$-statistic for
a null model with zero mean,
e.g. for $\Delta\Sigma$ we have
\begin{equation}\label{eq:ttest}
  t_{\Delta\Sigma} = \frac{\avg{\Delta\Sigma}}{\sigma_{\Delta\Sigma}/\sqrt{n}},
\end{equation}
where $\avg{\Delta\Sigma}$, $\sigma_{\Delta\Sigma}$ and $n$ are the
mean, standard deviation and size of the population; and analogously for
$t_{\Delta\text{AUC}}$. With a value of
$t_{\Delta\Sigma/\Delta\text{AUC}}$ and the sample size, we can obtain
the associated $p$-value, from which the null hypothesis can be rejected
if it is sufficiently small.

Figure~\ref{fig:auc_mdl} shows the results from model class comparisons
for all datasets in table~\ref{tab:data}.  Note that while the majority
of comparisons (81\%) are consistent, we observe a non-negligible
fraction of significantly inconsistent comparisons (19\%). Taking into
account the synthetic example from the previous section, the observed fraction of
inconsistent comparisons should not come as a surprise. Nevertheless, we
do not claim that the reason for the discrepancies observed in the
empirical data is precisely the same as the one in the planted partition
example.

\begin{table}
\begin{tabular}{l|r|r}
  Dataset & $N$ & $\avg{k}$ \\ \hline\hline
  American college football~\cite{girvan_community_2002} &$115$ &$10.7$\\
  Florida food web (dry)~\cite{ulanowicz_network_2005} &$128$ &$33.4$  \\
  Residence hall friendships~\cite{freeman_exploring_1998} &$217$ &$24.6$ \\
  \emph{C. elegans} neural network~\cite{white_structure_1986} &$297$ &$15.9$ \\
  Scientific coauthorships~\cite{newman_modularity_2006} &$379$ &$4.8$\\
  E-mail~\cite{guimera_self-similar_2003} &$1,133$ &$9.6$ \\
  Political blogs~\cite{adamic_political_2005} &$1,222$ &$31.2$ \\
  Crimes in St. Louis~\cite{kunegis_konect:_2013} &$1,380$ &$2.13$\\
  Protein iteractions (I)~\cite{stelzl_human_2005} &$1,706$ &$7.3$\\
  Bible name co-ocurrences~\cite{kunegis_konect:_2013} &$1,773$ &$10.3$\\
  Hamsterster  friendships~\cite{kunegis_konect:_2013} &$1,858$ &$13.5$\\
  Movielens ratings~\cite{kunegis_konect:_2013} &$2,625$ & $75.2$\\
  Adolescent friendships~\cite{moody_peer_2001} & $2,539$ & $10.2$\\
  Global airport network~\cite{peixoto_hierarchical_2014} &$3,286$ &$41.6$\\
  Protein iteractions (II)~\cite{joshi-tope_reactome:_2005} &$6,327$ &$46.6$\\
  Internet AS~\cite{leskovec_graph_2007} &$6,474$ &$4.3$\\
  Advogato user trust~\cite{massa_bowling_2009} &$6,541$ &$15.6$ \\
  Cora citations~\cite{subelj_model_2013} &$23,166$ &$7.9$ \\
  DBLP citations~\cite{ley_dblp_2002} & $12,591$ & $7.9$ \\
  Google+ social network~\cite{leskovec_learning_2012} &$23,628$ &$3.3$\\
  arXiv hep-th citations~\cite{leskovec_graph_2007} &$27,770$ &$25.4$\\
  Digg online converstations~\cite{choudhury_social_2009}&$30,398$ & $5.77$\\
  Linux source dependency~\cite{kunegis_konect:_2013} &$30,837$ &$13.9$\\
  PGP web of trust~\cite{richters_trust_2011} &$39,796$ &$15.2$ \\
  Facebook wall posts~\cite{viswanath_evolution_2009} &$46,952$ &$37.4$
\end{tabular}
\caption{Empirical networks used in this work, with their number of nodes
$N$ and average degree $\avg{k}=2E/N$.\label{tab:data}}
\end{table}

\subsection{Ensembles of simple models are more predictive than the single most compressive model}\label{sec:averaging}

We have shown that the model that best performs at link prediction is
often the most likely one or, equivalently, the one that best compresses
the data. Importantly, even for the cases in which both model selection
approaches are consistent, the single most compressive does not
necessarily provide optimal link predictions.

Indeed, according to Eq.~\ref{eq:linkpred_cond}, the best
approximation to the probability of a link is given by the average over
all partitions in a model class~\cite{hoeting_bayesian_1999}. Although
this average cannot be calculated exactly because of the combinatorially
large number of partitions, one can use Markov chain Monte Carlo (MCMC)
to sample over the partitions with the appropriate posterior
distribution $P(\bb|\Ao, \cl)$ \cite{guimera_missing_2009}. Then, the
average over all partitions is approximated by the average over the
sampled partitions, which asymptotically coincides with the exact value.

Note that if the posterior is dominated by a single partition (that is,
if the model is a perfect fit to the data) the single-point estimate
Eq.~\ref{eq:linkpred_mdl} will be an excellent approximation to the
average and these two approaches will coincide. However, when the model
is not a perfect fit, either due to lack of statistical evidence, or
more realistically, due to an imperfect description of the true
underlying generative mechanism, they will not.

\begin{figure*}
  \includegraphics[width=\textwidth]{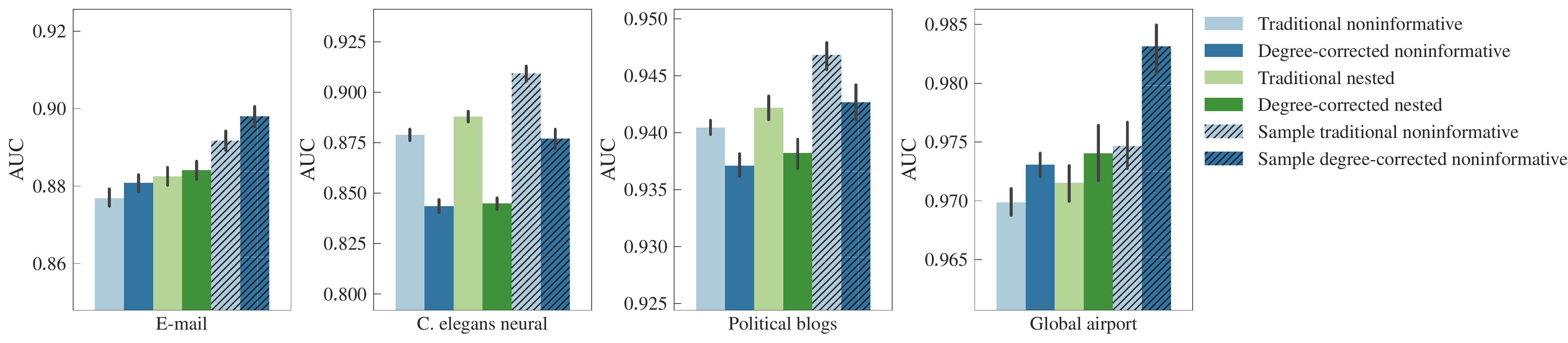} \caption{
  Comparison between single-point and average link prediction for four
  real networks and the four model classes. Single-point predictions are
  obtained using Eq.~\ref{eq:linkpred_mdl} whereas average predictions
  are obtained using MCMC sampling to approximate
  Eq.~\ref{eq:linkpred_cond}. In all cases, we removed a fraction
  $f=0.05$ of the true links of the network, and computed the AUC to
  measure predictive power (see text). Sampling with the simplest model
  class (the traditional non-informative SBM) always gives more accurate
  predictions than the single-point prediction with the best model.
  \label{fig:avg}}
\end{figure*}

For some of the networks in Table I, we have compared the predictive
power of single-point estimates with the four model classes, and
compared them to the predictive power of averages obtained using MCMC
sampling on the model classes with non-informative priors.\footnote{MCMC
sampling with the nested model classes is possible but too
computationally demanding for some of the networks
considered.} Fig.~\ref{fig:avg} shows that averaging over many
partitions improves the capacity of predicting missing edges, indicating
that the data is not perfectly described by the best
partition. Interestingly, the difference in AUC scores is often larger
between the best partition and the model average than it is across model
classes. This indicates that in-class variability is often more
expressive of the data (at least with respect to predictive performance)
than the single best fit of the most compressive model
class. Nevertheless, we still observe that the most compressive model
class also tends to yield higher predictive performance when averaged
over partitions\footnote{In order to properly extend our consistency
analysis in this case, instead of using single-point estimates, the
unsupervised approach would need to be based on the model evidence
$P(\A|\cl) = \sum_{\bb} P(\A|\bb,\cl)P(\bb)$, which then could be used
to compute the posterior odds ratio, as detailed in
Ref.~\cite{peixoto_nonparametric_2017}. Unfortunately, this quantity
cannot be computed in an asymptotically exact manner, even using MCMC.}.

\section{Conclusion}\label{sec:conclusion}

We have compared two approaches to model selection, one based on maximum
posterior likelihood (or maximum compression), and another based on
maximum performance at missing link prediction. We have found that while
these criteria tend to agree in practice, they fail to give consistent
results in some cases. In particular, we have seen that link prediction
can lead to overfitting because, perhaps counter-intuitively, overly
complex models sometimes give better predictions.

The fact that data prediction (in particular leave-one-out cross
validation) does not yield a consistent estimator of the underlying
generative process is well understood for linear
models~\cite{shao_linear_1993}, which is the same problem we have
observed for the SBM when only one link is removed. However, it was also
shown in Ref.~\cite{shao_linear_1993} that cross validation for linear
models is consistent if one performs leave-$k$-out, with $k$ scaling
proportionally with the number of data points in the training
set. However, doing so when the total amount of data is fixed, means we
must leave a large amount of data out of the inference procedure,
incurring a substantial loss of precision. We are thus left with two
competing goals: Increase the training set to maximize inference
precision, and increase the validation set to guarantee
consistency. Both can be achieved simultaneously only when the data is
plentiful, and when the model is well specified --- conditions that
cannot be always guaranteed in practice. For networks, even under the
SBM assumption, there are currently no good recipes to reach the proper
balance, or an assurance that such a balance even exists. This problem
is particularly exacerbated by the fact that most networks are sparse,
and hence there might be insufficient data to confidently identify the
correct model even when they are infinitely large. In fact, a removal of
a fraction of edges will always make the network sparser, potentially
crossing a threshold that makes the latent structure completely
undetectable~\cite{decelle_asymptotic_2011}.

An important ramification of our results is that the potential overfitting that can arise out of
seeking the best predictions does not mean that one should avoid doing
it altogether. On the contrary, overfitting becomes a non-issue if the main objective is to generalize
from previous observations and guess possible errors and omissions in
the data, or predict future observations, with the highest precision. In this situation we have shown that
the best approach is, in fact, to average over models, rather than use
a single model. In any case, one should always be careful not to
conclude that the preferred model or models in this situation are
closer to the actual underlying generative process.

\begin{acknowledgments}
This work was supported by a James S. McDonnell Foundation Research
Award (M.S.-P. and R.G.), and by Grants FIS2016-78904-C3-1-P
(M.S.-P. and R.G.) and FIS2015-71563-ERC (to R.G.) from the Spanish
Ministerio de Econom\'{\i}a y Comptetitividad.
\end{acknowledgments}

\bibliography{bib}

% ---------------------------------------------------

\appendix
\section{Posterior probability of missing links and non-links}
\label{app:link_pred}

Our goal is to obtain an expression for the posterior likelihood of
missing entries $P(\dA|\Ao)$, conditioned on the observed network
$\Ao$. We will make use of only two simple assumptions about the
data-generating process. First, we assume that the complete network
$\G=\Ao\cup\dA$ is sampled from some version of the SBM with a marginal
likelihood
\begin{equation*}
  P_G(\G|\bb) = P(\G|\bb,\cl).
\end{equation*}
Secondly, given a generated network $\G$, we then select a portion of
the entries $\dA$ from some distribution,
\begin{align}\label{eq:PdA}
  P_{\delta A}(\dA|\G),
\end{align}
which models our source of errors.  The observed network is obtained
from the above construction by removing $\dA$ from $\G$,
\begin{equation}\label{eq:cons}
  \Ao = \G\setminus\dA,
\end{equation}
where the notation above means that the edges and non-edges present in
$\dA$ and $\G$ are left indeterminate in $\Ao$ (although,
e.g. considering edges them as non-edges in $\Ao$, i.e. $\Ao=\G-\dA$
would yield an identical outcome, as we show below). Given the above
model, we want to write down the joint likelihood $P(\Ao,\dA|\bb)$, so
that we can obtain the conditional likelihood $P(\dA|\Ao,\bb)$. We begin
by using Eq.~\ref{eq:cons} to write
\begin{align*}
  P(\Ao|\dA, \G) &= \delta(\Ao - (\G \setminus \dA)), \\
  &=  \delta(\G - (\Ao\cup\dA)),
\end{align*}
since there is only one possibility which is consistent, where
$\delta(\bm{B}-\bm{C}) = 1$ if $\bm{B} = \bm{C}$ or $0$ otherwise.
Thus, if we know the complete graph $\G$, we can write the joint
likelihood as
\begin{align*}
  P(\Ao,\dA|\G) &= P(\Ao|\dA, \G) P_{\delta A}(\dA|\G),\\
                &=  \delta(\G - (\Ao\cup\dA))P_{\delta A}(\dA|\G).
\end{align*}
Finally, for the joint distribution conditioned on the partition, we sum
the above over all possible graphs $\G$, sampled from our original model,
\begin{align*}
  P(\Ao,\dA|\bb) &= \sum_{\G}P(\Ao,\dA|\G)P_G(\G|\bb)\\
                 &= P_{\delta A}(\dA|\Ao\cup\dA)P_G(\Ao\cup\dA|\bb).
\end{align*}
From this, we can write directly our desired posterior of missing entries
by averaging over all possible partitions,
\begin{align}
  P(\dA|\Ao) &= \frac{\sum_{\bb}P(\Ao,\dA|\bb)P(\bb)}{P(\Ao)} \\
  &= \frac{P_{\delta A}(\dA|\Ao\cup\dA)\sum_{\bb}P_G(\Ao\cup\dA|\bb)P(\bb)}{P(\Ao)},\label{eq:cond_dA}
\end{align}
with $P(\Ao)$ being a normalization constant, independent of $\dA$.
Note that the equation above does not depend on whether $\Ao$ includes
the missing entries as edges or non-edges, or if they are left
indeterminate as we have, as the only relevant quantity in the numerator
is the complete graph $\Ao\cup \dA$. Therefore, even though these
representations amount to very different interpretations of the data,
they result in the same inference outcome, since in the end all that
matters is the model we have for the complete network.

Although it is complete, Eq.~\ref{eq:cond_dA} cannot be used directly to
compute posterior likelihood, as it includes a sum over all
partitions. It does, however, suggest a simple algorithm: We could
compute the average of $P_G(\Ao\cup\dA|\bb)$ by sampling many partitions
$\bb$ from the prior $P(\bb)$. However, even though it is correct, this
algorithm will typically take an astronomical time to converge to the
asymptotic value, since the largest values of $P_G(\Ao\cup\dA|\bb)$ will
be far away from the typical values of $\bb$ sampled from
$P(\bb)$. Instead, a much better algorithm is obtained by performing
importance sampling, i.e. by writing the likelihood as
\begin{widetext}
\begin{align}
  P(\dA|\Ao) &\propto P_{\delta A}(\dA|\Ao\cup\dA)\sum_{\bb}P_G(\Ao\cup\dA|\bb)\frac{P_G(\Ao|\bb)}{P_G(\Ao|\bb)}P(\bb),\nonumber\\
  &\propto P_{\delta A}(\dA|\Ao\cup\dA)\sum_{\bb}\frac{P_G(\Ao\cup\dA|\bb)}{P_G(\Ao|\bb)}P_G(\bb|\Ao).
\end{align}
\end{widetext}
where we have used
\begin{align*}
  P_G(\bb|\Ao) = \frac{P_G(\Ao|\bb)P(\bb)}{P_G(\Ao)},
\end{align*}
which is the posterior of $\bb$ pretending that $\Ao$ came directly from
the SBM, which we can sample efficiently using MCMC. Naturally, if the
number of entries in $\dA$ is much smaller than in $\Ao$, this posterior
distribution will be much closer to the region of interest, and the
estimation of the likelihood will converge significantly faster. Note,
however, that in order to compute $P_G(\Ao|\bb)$ and sample from
$P_G(\bb|\Ao)$ we must decide whether the missing edges/non-edges in
$\Ao$ are really missing or if we replace them with zeros or ones. The
choice, however, cannot change the resulting distribution $P(\dA|\Ao)$,
as it is invariant with respect to the weights we use when doing
importance sampling. Hence, the choice we make should done purely on
algorithmic grounds. In our experiments we will consider missing edges
(non-edges) as non-edges (edges), since it allows MCMC implementations
developed for this case to be used without modification.

To complete the estimation, we need to define how the edges and
non-edges are removed from the original network. Without loss of
generality, focusing on the case of missing edges only, a simple
assumption is a uniform distribution conditioned on the fraction of
missing edges $f$,
\begin{align}\label{eq:PdAf}
  P_{\delta A}(\dA|\G,f) &= \prod_{i<j}{G_{ij}\choose \delta A_{ij}} f^{\delta A_{ij}}(1-f)^{G_{ij}-\delta A_{ij}}\nonumber\\
                    &= f^{E_\delta}(1-f)^{E_G-E_{\delta}},
\end{align}
where $E_\delta$ and $E_G$ are the total number of edges that are
removed and in the original network, respectively, and we have assumed a
simple graph in the last equation for simplicity. If we are always
considering the same number of missing edges, Eq.~\ref{eq:PdAf} is only a
constant, resulting in
\begin{equation}
  P(\dA|\Ao) \propto \sum_{\bb}\frac{P_G(\Ao\cup\dA|\bb)}{P_G(\Ao|\bb)}P_G(\bb|\Ao).\label{eq:posterior_dA_detailed}
\end{equation}
which is Eq.~\ref{eq:linkpred_cond} in the main text. This equation is
exact up to a normalization constant that is often unnecessary to
compute, as we are mostly interest in relative probabilities of missing
edges. We stress that in deriving Eq.~\ref{eq:posterior_dA_detailed} we
have not made any reference to the internal structure of the network
model $P_G(\G|\bb)$, and is equally valid not only for all model
variants used in this work, but also to a much wider class. This is in
contrast to similar frameworks that have been derived with much more
specific models in
mind~\cite{clauset_hierarchical_2008,guimera_missing_2009}. Furthermore,
we note also that although we have assumed in the last steps that $\dA$
is a set of missing edges, the same argument above can be adapted with
almost no changes when it represents instead any arbitrary combination
of missing and spurious edges, and hence our framework can be used in
this more general scenario as well.

We note that the problem of selecting the most appropriate fraction of
missing edges with the objective of performing model selection is not a
trivial one. In fact, only creating missing edges but not spurious ones
is a biased way to proceed, since a more accurate representation of the
data would consider edges and non-edges on equal footing. However,
choosing the optimal relative fraction would require not only preserving
the sparsity of the data (i.e. selecting a larger fraction of missing
edges than spurious ones) but also more information about the
heterogeneous mixture of edge populations, which would depend on the
true model parameters. We leave this open problem for a future work, and
concentrate instead of the more typical task of missing edge prediction.

\section{Link prediction is not always a good model selection criterion:
  The planted partition example}\label{app:auc}

\begin{figure*}
  \begin{tabular}{ccc}
    \begin{overpic}[width=.33\textwidth]{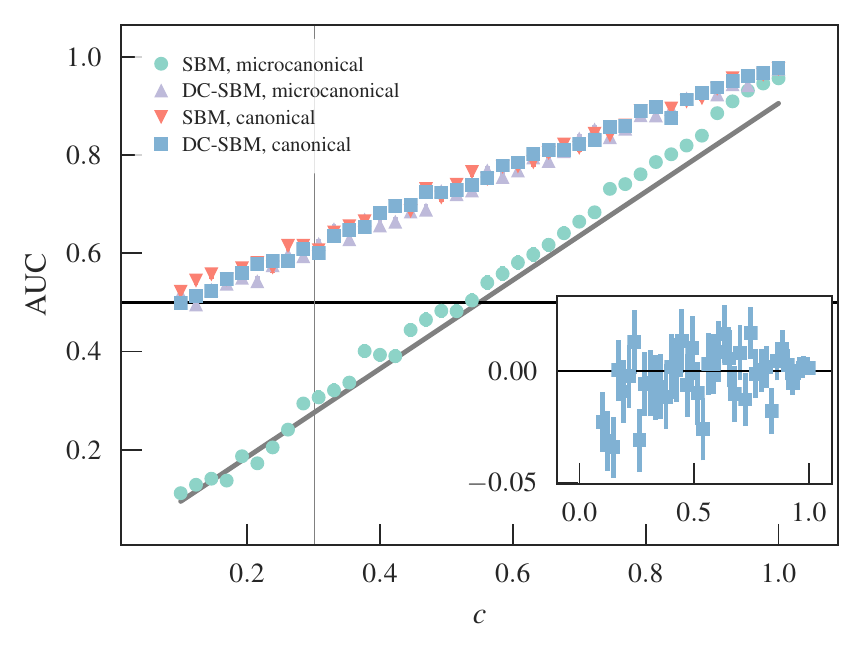}
      \put(0,10){(a)}
    \end{overpic}&
    \begin{overpic}[width=.33\textwidth]{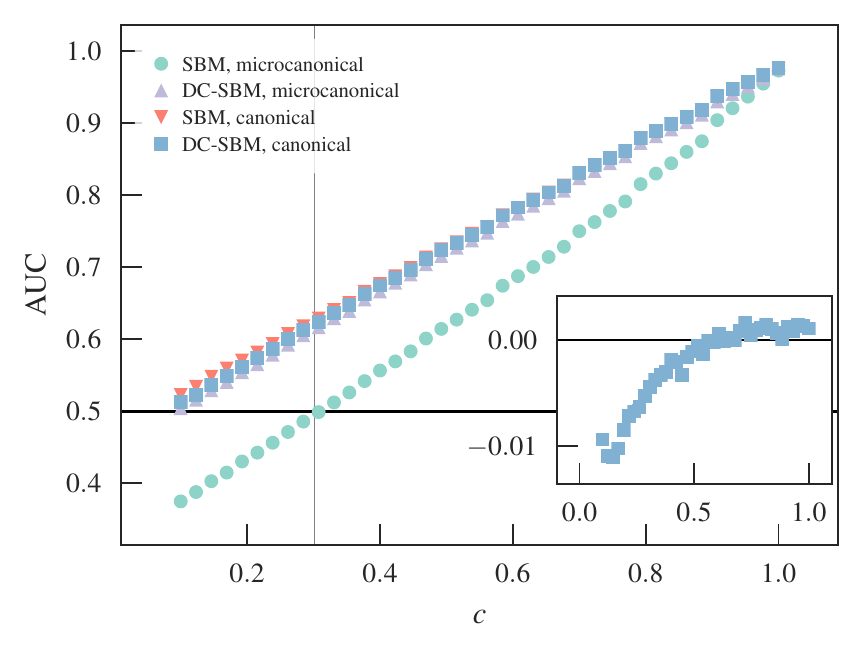}
      \put(0,10){(b)}
    \end{overpic}&
    \begin{overpic}[width=.33\textwidth]{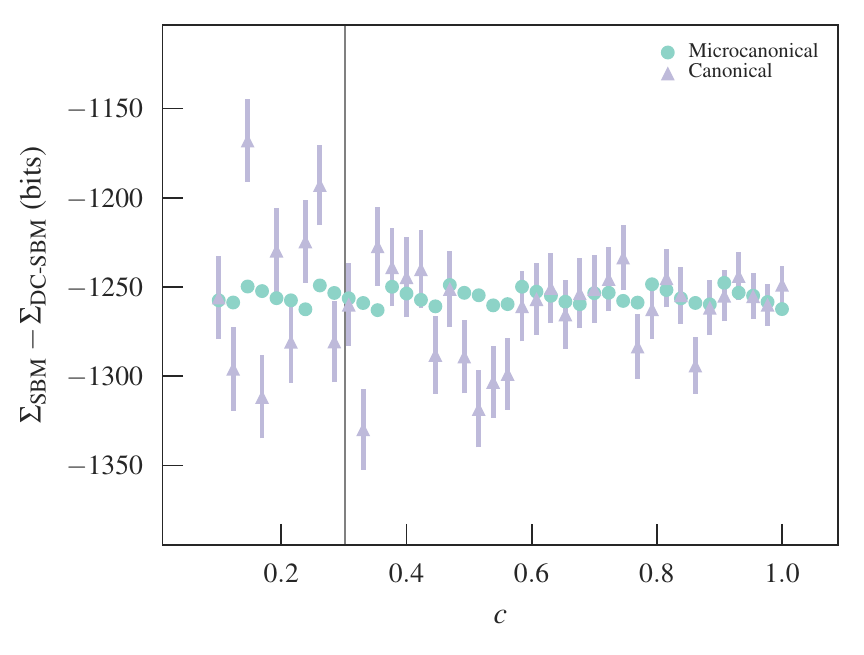}
      \put(0,10){(c)}
    \end{overpic}
  \end{tabular}

  \caption{(a) Average AUC values obtained by removals of a single edge
  from a PP with $n_r=100$, $B=10$ and $\avg{k}=20$, both for
  ``canonical'' (i.e. unconstrained) as well as microcanonical samples,
  where Eq.~\ref{eq:micro} holds. The legend indicates which model was
  used to compute the AUC (i.e. the SBM or the DC-SBM). The solid line
  corresponds to Eq.~\ref{eq:auc_pp}, and the vertical line the value
  $c^*=1/B+(B-1)/(B\sqrt{\avg{k}})$ corresponding to the detectability
  threshold.  The inset shows the difference of the AUC values obtained
  with the two model classes, $\text{AUC}_{\text{DC-SBM}} -
  \text{AUC}_{\text{SBM}}$, with networks sampled from the canonical
  model. (b) The same as (a), but with a fraction $f=0.05$ of the edges
  removed.
  (c) Description length difference between the SBM and DC-SBM, both for
  the canonical and microcanonical samples, for a fraction $f=0.05$ of
  the edges removed.\label{fig:loo}}
\end{figure*}

\begin{figure}
  \begin{tabular}{cc}
    \begin{overpic}[width=.49\columnwidth]{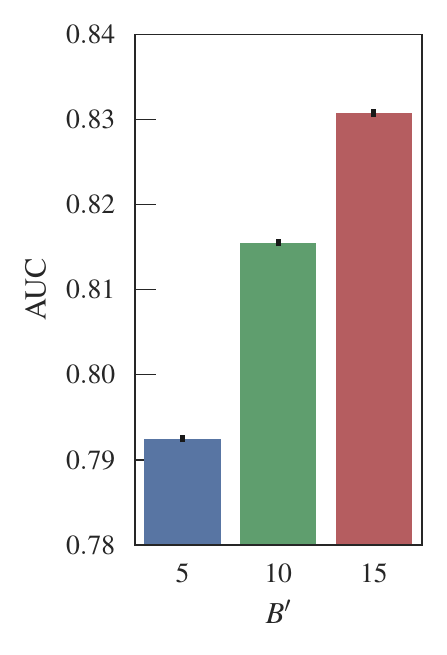}
      \put(0,10){(a)}
    \end{overpic}&
    \begin{overpic}[width=.49\columnwidth]{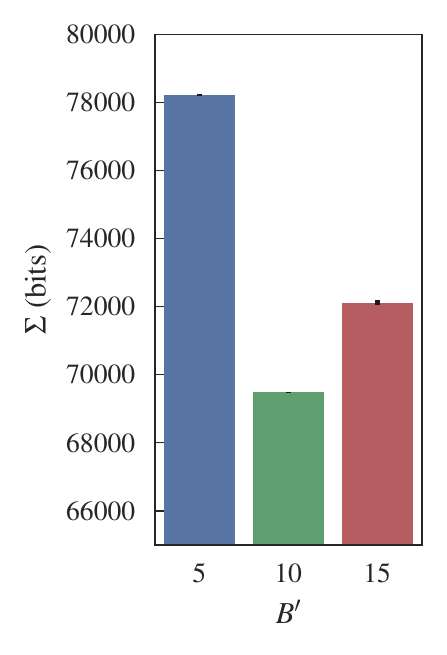}
      \put(0,10){(b)}
    \end{overpic}
  \end{tabular} \caption{(a) AUC values obtained by removing a fraction
  $f=0.05$ of the edges of a PP with $n_r=100$, $B=10$, $\avg{k}=20$ and
  $c=0.8$, for microcanonical samples fulfilling Eq.~\ref{eq:micro}, and
  forcing $B=B'$ during inference. (b) The same as (a), but with the
  description length $\Sigma$, instead. The MDL criterion pinpoints the
  correct planted value of $B'=10$, whereas AUC
  overfits. \label{fig:overfit}}
\end{figure}

We consider a simple parametrization of the non-degree-corrected SBM
known as the planted-partition model (PP), which is composed of $N$
nodes divided into $B$ equal-sized groups and is generated according to
Eq.~\ref{eq:poisson_sbm} with
\begin{equation}\label{eq:pp}
  \lambda_{rs} = \frac{2\avg{E}}{n_rn_s}\left[\frac{c\delta_{rs}}{B}+\frac{(1-c)(1-\delta_{rs})}{B(B-1)}\right],
\end{equation}
where $n_r = N/B$, $\avg{E}$ is the average number of edges, and
$c\in[0,1]$ controls the degree of assortativity between groups. For
$c>1/B$ the placement of edges is not fully random, and for
$c>1/B+(B-1)/(B\sqrt{\avg{k}})$ the planted modular structure is
detectable from the data alone~\cite{decelle_asymptotic_2011}. In the
following discussion we assume that $c>1/B$ and that the partition of
the nodes is always known a priori.

Specifically, we consider networks which have an observed number of edges
between groups that matches exactly the expected value,
\begin{equation}\label{eq:micro}
  e_{rs} = \sum_{ij}A_{ij}\delta_{b_ir}\delta_{b_is} = \lfloor n_rn_s\lambda_{rs}\rceil,
\end{equation}
where $\lfloor x\rceil$ rounds $x$ to the nearest integer.

When faced with an instance of this model, we want to
evaluate the predictiveness of the model by performing leave-one-out
cross validation: We remove a single edge from the network, and consider
its likelihood according to the observed network. Based on this we
compute the AUC, i.e. the probability that the removed edge is ranked
above the false positives. Here we show how the result of this
experiment can be computed analytically.

We begin by considering a slightly different scenario: Instead of
computing the likelihood of the missing edge via the posterior
distribution, we use instead the true likelihood of the original model,
before we removed the edge. When doing so, because of the symmetries in
the model, there will be only two possible values of the likelihood,
depending only on whether the removed edge lies between nodes of the
same or different groups. If the removed edge connects nodes of the same
group, the only false positives that have the same likelihood will be
those that also connect nodes of the same group (although they do not
need to be the same group of the removed edge), and the remaining edges
will have a lower likelihood. With this, and assuming that $N\gg 1$ and
sufficiently sparse networks so that $e_{rs}<n_rn_s$, the computed AUC
will be
\begin{align}
  \text{AUC}_{\text{in}} &= \frac{1}{N^2/2}\left[\frac{1}{2}\frac{Bn_r^2}{2}+\frac{B(B-1)}{2}n_r^2\right]\\
   &= \frac{1}{2B} + \frac{(B-1)}{B},
\end{align}
which means we have $\text{AUC}_{\text{in}} > 1/2$ if $B>1$, indicating
that we can predict the missing edge better than pure chance. For
removed edges between different groups, we have instead,
\begin{align}
  \text{AUC}_{\text{out}} &= \frac{1}{N^2/2}\frac{1}{2}\frac{B(B-1)}{2}n_r^2\\
   &= \frac{(B-1)}{2B},
\end{align}
from which we see that $\text{AUC}_{\text{out}} < 1/2$, i.e. edges
between groups are predicted with a performance that is inferior to
fully random guesses. Overall, the average performance for randomly chosen
edges is
\begin{align}
  \text{AUC} &= c\text{AUC}_{\text{in}} + (1-c)\text{AUC}_{\text{out}} \\
             &= c\left[\frac{1}{2B}+\frac{(B-1)}{2B}\right] + \frac{B-1}{2B}.
\end{align}
For any $c>1/B$ we have that $\text{AUC}>1/2$, meaning that the
generative model on average provides better predictions of randomly
missing edges that are better than pure chance. This behavior is fully
expected, since the process generating the missing edges is not random,
and is described precisely by our model.

However, in the scenario of an actual missing link, we need to infer the
model from the observed data, in the absence of the
removed edge. If the removed edge connects groups $r$ and $s$, the new
edge counts between these two groups will be $(e_{rs}-1 -\delta_{rs})$,
and hence the posterior likelihood of observing a missing link there
will be slightly smaller than in the true model. Since in the original
model all other edges of the same kind (inter or intra-group) had
exactly the same likelihood, this small difference in the likelihood
will be sufficient to make the actual missing edge less likely than all
the other ones with the same likelihood originally. Because of this, in
this situation we have
\begin{align}
  \text{AUC}_{\text{in}} &= \frac{1}{N^2/2}\left[\frac{1}{2}\frac{n_r^2}{2}+\frac{B(B-1)}{2}n_r^2\right]\\
   &= \frac{1}{2B^2} + \frac{(B-1)}{B},
\end{align}
and
\begin{align}
  \text{AUC}_{\text{out}} &= \frac{1}{N^2/2}\frac{1}{2}n_r^2\\
   &= \frac{1}{2B^2},
\end{align}
and thus
\begin{align}\label{eq:auc_pp}
  \text{AUC} &= c\text{AUC}_{\text{in}} + (1-c)\text{AUC}_{\text{out}} \\
             &= \frac{1}{2B^2}+c\frac{B-1}{B}.
\end{align}
Differently from the case where the true model is known, now if $1/B < c
< (B^2-1)/[2B(B-1)]$ we have a non-random inferred model that yields
$\text{AUC} < 1/2$, and thus an inferior predictive performance than
pure chance, despite the fact that the model differs from the true one
only minimally. The reason for this is that the removal of \emph{any
single} edge decreases its probability --- according to the model
inferred from the remaining network --- below a vast number of false
positives (i.e. edges of the same kind), which in fact have the exact
same likelihood under the original model.

As mentioned in the main text, if we infer using the wrong model class,
for example the DC-SBM, we systematically observe larger AUC values, as
can be seen in Fig.~\ref{fig:loo}a. This is because the extra parameters
of this model
--- the degree propensities $\theta_i$ --- incorporate a
large amount of noise from the data and destroy the homogeneity present
in the simpler model. Without the homogeneity, the single edge count
lost between groups $r$ and $s$ makes little difference overall. As can
also be seen in Fig.~\ref{fig:loo}b, this phenomenon persists even if we
remove a finite fraction of the edges, instead of a single one.

Despite the improved predictive performance, the DC-SBM is not the most
appropriate model for this network. Not only we generated the data
explicitly from the simpler SBM, but also its posterior likelihood is
smaller, as reflected by its larger description length (see
Fig.~\ref{fig:loo}c). Hence, the unsupervised model selection approach is
impervious to details of the model such as the fact that the edge
probabilities are similar, and correctly identifies the true generative
process. We emphasize that even if one would stubbornly prefer the most
predictive model in this case, one would have to accept a fully random
network over the simpler SBM, when the later yields AUC values smaller
than 1/2.

The reason why link prediction fails to select the true underlying model
in this case is not the lack of statistical evidence, but rather that
the model itself --- and not the data --- is sensitive to perturbations:
A minimal change to one of the $\lambda_{rs}$ values downgrades or
upgrades the likelihood of the respective edges with respect to all
others of different types that would otherwise have the exact same
probability. Hence, this example illustrates how in some cases
predictive performance (at least when measured by the AUC) can be to
some extent an inherent property of a model, regardless of its quality
of fit to the data.

One could argue that although the networks that obey Eq.~\ref{eq:micro}
have the largest probability, they are nevertheless not representative
of the whole ensemble: Since the edge counts $e_{rs}$ are sums of
Poisson variables, they are also distributed according to a Poisson, and
therefore their probabilities of matching exactly the expected values
will decrease as $P(e_{rs}=n_rn_s\lambda_{rs}) \approx
1/\sqrt{n_rn_s\lambda_{rs}}$, for large arguments. For large and sparse
networks, this value will decrease as $1/\sqrt{N}$, and hence, despite
being the most likely type of network, its absolute probability will be
very small asymptotically, and therefore most networks sampled from this
model will not possess such an extreme level of homogeneity. Because of
this, one could say that this is an ``out-of-class'' example, and that
would perhaps explain the inconsistency.  Although this is technically
true, it is easy to see that this argument is a red herring: We can
easily view the above case as a typical instance of an equivalent
\emph{microcanonical} model~\cite{peixoto_nonparametric_2017}, where the
homogeneity of Eq.~\ref{eq:micro} is strictly imposed for all sampled
networks, and the rest of the analysis would still remain
valid. Nevertheless, we can also show that the same problem occurs for
typical samples from the original ensemble, which do not necessarily
conform to Eq.~\ref{eq:micro}, albeit less prominently. As seen in the
inset of Fig.~\ref{fig:loo}b, for a range of the parameter $c$ --- in
particular when the structure of the model is strongest --- we still
observe higher AUC values for the DC-SBM, at least when the fraction of
removed edges is sufficiently large. The explanation we offer is the
same: the fluctuations are not always sufficient to mask the homogeneity
in the true model, which thwarts the predictability of missing edges.

The above phenomenon also interferes with the selection of the number of
groups. Link prediction has been proposed before as a means of selecting
the number of groups~\cite{kawamoto_cross-validation_2016}, as well as
other dimensional aspects~\cite{de_bacco_community_2017}, but as we show
in Fig.~\ref{fig:overfit} it also fails for precisely the same reason:
increasing the number of groups incorporates more noise in the model,
and breaks its homogeneity. This leads to a clear overfitting, where
spurious groups are identified. As before, unsupervised model selection
is not susceptible to this, and reliably selects the correct number of
groups. Because of this possibility, we admonish against using the
supervised approach in favor of the unsupervised for this purpose.

\end{document}